%% file: Mawdoo3_DeepQ2Q_ArxiV.tex
\documentclass{article}

\usepackage{arabtex}
\usepackage{arxiv}

\usepackage[utf8]{inputenc} % allow utf-8 input
\usepackage[T1]{fontenc}    % use 8-bit T1 fonts
\usepackage{hyperref}       % hyperlinks
\usepackage{url}            % simple URL typesetting
\usepackage{booktabs}       % professional-quality tables
\usepackage{amsfonts}       % blackboard math symbols
\usepackage{nicefrac}       % compact symbols for 1/2, etc.
\usepackage{microtype}      % microtypography
\usepackage{lipsum}
\usepackage{cite}
\usepackage{mathtools}
\DeclarePairedDelimiterX{\inp}[2]{\langle}{\rangle}{#1, #2}
\usepackage{amsmath,amssymb,amsfonts}
\usepackage{algorithmic}
\usepackage{graphicx}
\usepackage{textcomp}
\usepackage{xcolor}
\usepackage{soul} % could be removed 
\title{Deep Contextualized Pairwise Semantic Similarity for Arabic Language Questions}

\author{
  Hesham Al-Bataineh %\thanks{Use footnote for providing further    information about author (webpage, alternative    address)---\emph{not} for acknowledging funding agencies.} 
  \\
  AI Department\\
  Mawdoo3 Ltd\\
  Amman, Jordan \\
  \texttt{hisham.bataineh@mawdoo3.com} \\
   \And
 Wael Farhan \\
  AI Department\\
  Mawdoo3 Ltd\\
  Amman, Jordan \\
  \texttt{wael.farhan@mawdoo3.com} \\
   \AND
   Ahmad Mustafa \\
   AI Department\\
   Mawdoo3 Ltd \\
   Amman, Jordan \\
   \texttt{ahmad.mustafa@mawdoo3.com} \\
   \And
   Haitham Seelawi\\
   \texttt{haitham.seelawi@gmail.com} \\
   \And
   Hussein T.~Al-Natsheh \\
   AI Department\\
   Mawdoo3 Ltd \\
   Amman, Jordan \\
   \texttt{h.natsheh@mawdoo3.com} \\
}

\begin{document}
\maketitle

\begin{abstract}
\input{sections/abstract.tex} 
\end{abstract}

% keywords can be removed
\keywords{Arabic, natural language processing , semantic textual similarity, question answering, deep learning}

\section{Introduction}
\label{sec:introduction}
\input{sections/introduction.tex}

\section{Related Work}
\label{sec:related}
\input{sections/related.tex}

\section{Data}
\label{sec:data}
\input{sections/data.tex}

\section{Models}
\label{sec:models}
\input{sections/models.tex}

\section{Experimental Setup}
\label{sec:setup}
\input{sections/experimental.tex}

\section{Conclusion and Future Work}
\label{sec:conclusion}
\input{sections/conclusion.tex}
\clearpage
\bibliographystyle{unsrt}  
\bibliography{references}  %%% Remove comment to use the external .bib file (using bibtex).
%%% and comment out the ``thebibliography'' section.

%%% Comment out this section when you \bibliography{references} is enabled.
%\begin{thebibliography}{1}

\end{document}

%% file: sections/abstract.tex
Question semantic similarity is a challenging and active research problem that is very useful in many NLP applications, such as detecting duplicate questions in community question answering platforms such as Quora. Arabic is considered to be an under-resourced language, has many dialects, and rich in  morphology. Combined together, these challenges make identifying semantically similar questions in Arabic even more difficult. In this paper, we introduce a novel approach to tackle this problem, and test it on two benchmarks; one for Modern Standard Arabic (MSA), and another for the 24 major Arabic dialects.
We are able to show that our new system outperforms state-of-the-art approaches by achieving 93\% F1-score on the MSA benchmark and 82\% on the dialectical one. 
This is achieved by utilizing contextualized word representations (ELMo embeddings) trained on a text corpus containing MSA and dialectic sentences. 
This in combination with a pairwise fine-grained similarity layer, helps our question-to-question similarity model to generalize predictions on different dialects while being trained only on question-to-question MSA data.

%% file: sections/introduction.tex
% Hussein
A question is a linguistic form that does not necessarily provide information as in a sentence but may carry semantic information. 
The closest level of granularity to the question form is the sentence form rather than word or paragraph. 
The answer, on the other hand, could be mapped either to a sentence or a paragraph as the closest granularity. 
In this work, we are focusing one Semantic Textual Similarity (STS) of a question pair with the main assumption that if two questions have the same answers, then they are semantically equivalent. 
Otherwise, if the answers are different or partially different (e.g., the answer of the first question is included in the answer of the second question but considered incomplete, or the questions share a portion of their answers yet different otherwise) then the pair is considered as semantically nonequivalent.

STS deals with resolving semantic similarity between the different levels of expression in a given language (e.g. words, sentences, paragraphs). 
At heart, STS is a Natural Language Understanding (NLU). With the advent of deep learning methods and its applications in NLU, STS has especially witnessed huge advances, such as reported by \cite{yang2018learning} and \cite{cer2018universal}. 
These advances have resulted in the development of complex applications in the field that previously were thought very difficult to achieve. 
Nonetheless, it seems that STS in Arabic is yet to reap these benefits. 
Despite a plethora of literature reporting Arabic STS systems developed using conventional machine learning methods \cite{alian2018arabic}, only a few of them utilize dense word representations in STS related tasks, let alone deep learning \cite{al2018deep}.

% Hussein
Our interest in tackling this problem stems from the fact that one of the most relevant use-cases of STS for question pairs is identifying duplicate questions in community question answering platforms. In such platforms, the user provides a question that might be already asked and answered by the community but the question might be phrased differently. Looking for an exact match, even after text normalization, would not solve the problem and would result in information duplication and a waste of efforts in answering already answered questions. It is an intellectual task that a human would best decide if two questions, or more, are semantically equivalent. 
For example, the following pair of questions; ``How to prepare scrambled eggs?'' and `` What are the steps of making good scrambled eggs?'' could have the same answer, however, a text matching approach would not identify that they are semantically similar. 
Search engines of a question repository, for example, FAQ, is another application where such a question-to-question (Q2Q) similarity model would be very useful. 
Digital assistants like chat-bots or voice assistants also need such NLU model to map the user question to already indexed answered questions.

Therefore, it is in this vein that we produce and publish this work; in it, we attempt to develop state of the art Arabic Q2Q systems, that utilize recent advances in the field. By this, we hope to encourage other members of the Arabic Computational Linguistics community to open up to and embrace the game-changing methodologies of deep learning, and its potential to revolutionize Arabic NLP and NLU.

Working in Arabic Q2Q similarity can be a challenging task due to several factors:
\begin{enumerate}
    \item \textbf{Under-resourced:} Arabic Q2Q datasets are scarce and limited in size.
    \item \textbf{Dialectic:} Modern Standard Arabic (MSA) is not commonly used in social media, as most users prefer using their own dialect (a variation of the language). Where each dialect has its own vocabulary and mutations of existing words.
    Training models for dialects can be problematic for two main reasons. First, there is not enough labeled dataset available for each dialect. e.g., Quora Question Pair\footnote{https://data.quora.com/First-Quora-Dataset-Release-Question-Pairs} and other datasets support standard English, but very few support Scottish, Irish, and other dialects. This applies to most languages including Arabic. Secondly, if labeled dialectic data is abundant then it would be tedious to train one model per dialect, where each model cannot jointly find similar pairs between different dialects.
    \item \textbf{Morphologically rich:} The Arabic language is one of the most morphologically rich languages which complicates training traditional word embedding algorithms such as Word2vec as it needs to learn a completely new embedding for each morphology.
\end{enumerate}

The main purpose of this paper is to tackle those issues and set a new state-of-the-art Q2Q similarity score for the Arabic language. Our contributions can be summarized into the following points:
\begin{enumerate}
    \item Build a new Q2Q standard Arabic dataset \cite{seelawi-EtAl:2019:NSURL} that is publicly available and large enough to train deep learning models. Currently, this is the largest Arabic Q2Q dataset.
    \item We extract a Q2Q evaluation dataset from MADAR Arabic dialect parallel corpora \cite{habash2018unified}. This second test set is created to evaluate our models on how they generalize on different dialects.
    \item Train a new ELMo model that starts with a character-level convolutional neural network (CNN) to overcome the hurdle of morphology and out-of-vocabulary (OOV). The training corpus was collected from various sources that include dialectic data. Our word embeddings model achieves better results than the best known Arabic word embeddings AraVec \cite{aravec}
    \item We build a deep learning network that is only trained on MSA without including any dialectic question pairs in the training set. Then, we show that this network is able to generalize and achieve high F1 score on the dialectic test set.
\end{enumerate}

In section~\ref{sec:related}, we discuss previously published work relating to Q2Q in Arabic. We confine our review to research that utilizes the recent advances in neural networks. Section \ref{sec:data} describes datasets collected to train our models, while section \ref{sec:models} dives into our approaches. Next, we describe our experiments and analyze the results in section \ref{sec:setup}. Finally, we conclude in section \ref{sec:conclusion}.

%% file: sections/related.tex
%Hussein
There is not much work on Arabic Q2Q semantic similarity in particular. However, there are a few in Arabic sentence similarity (e.g., STS) and a lot in other languages, mainly in English. In this section, we will provide a related work review of sentence STS similarity approaches categorized into three main approaches, unsupervised models, deep learning models, and computational linguistics approach.

% \subsection{Word Embedding-Based Models}
\input{sections/related/unsupervised.tex}

% \subsection{Deep Learning Models}
\input{sections/related/deep_learning.tex}

%Hussein
% \subsection{Computational Linguistics Approaches}
\input{sections/related/CL.tex}

%% file: sections/related/unsupervised.tex
In the unsupervised models, Nagoudi et al. \cite{schwab2017semantic} are among the first to use dense word representations to tackle STS in Arabic.
They propose the usage of Word2vec embeddings \cite{mikolov2013efficient}, by carrying a point-wise summation of the dense vector representation of each constituent word for each sentence in a given pair.
They then use cosine similarity on the resultant vector representations to score how semantically similar or not the two sentences are.
They have used the CBOW variant of the Word2vec model proposed in \cite{zahran2015word}, which is trained on an Arabic corpus of about 5.8 billion tokens. 

They have experimented with weighing the summations using tf-idf and part-of-speech (POS) tagging, in juxtaposition with no weighting at all. 
The testing dataset they have used consists of 750 sentences from the MSR-Video corpus, which was manually translated to Arabic.
They have found that the two weighting schemes ($78.20\%$ Pearson correlation for tf-idf, and $79.69\%$ for POS tagging) significantly outperforms not using weighting at all ($72.33\%$). 
This is expected, given that the words making up a given sentence do not contribute equally to its semantics.
However, while tf-idf weights are automatically chosen from the training text, assigning weights to POS tags requires domain knowledge.
In our approach, we achieve better results without requiring domain knowledge.
% For the sake of completeness, they have used a CBOW variant of the Word2vec model \cite{zahran2015word} trained on an Arabic corpus of about 5.8 billion tokens. 
% For the test dataset upon which the above-mentioned performance was obtained, they have used 750 sentences from the MSR-Video corpus, which was manually translated to Arabic.

%% file: sections/related/deep_learning.tex
As mentioned in the introduction, there is a lack of deep learning studies on the topic of STS in Arabic. 
As part of their efforts for SemEval-2017 Task 1, \cite{cer2017semeval} and \cite{shao2017hcti} successfully but \textit{indirectly} use a standalone, end-to-end deep learning solution for detecting semantic similarity between pairs of Arabic sentences. 
By indirectly, we mean that they train the system on English pairs, and then for inference, they translate the pairs from Arabic to English using Google Translate service.
% to obtain the scores.

The architecture of the system itself is heavily inspired by the Convolutional Deep Structured Semantic Model as suggested by \cite{huang2013learning}. 
The model uses Glove embeddings \cite{pennington2014glove} to represent words. 
Additionally, the POS tag is appended, as a one-hot encoding vector, to the Glove representation of its corresponding token.
As such, each sentence is modeled as a matrix of dense representations (for words) plus sparse representations (for POS), arranged in a manner corresponding to the order of the words in the original sentence. 
The matrix is then padded/truncated to a maximum sentence length.

The matrices of a pair of sentences are then processed using a convolutional neural network \cite{lecun1999object} with 300 one-dimensional filters, and a max-pooling step \cite{scherer2010evaluation}. 
The resultant sentence vector representations are then subjected to two more transformations: an absolute point-wise difference, and a point-wise multiplication, both, between the two said vectors.
These are then concatenated and fed into a feed-forward network, which predicts the similarity between two sentences on a scale between 1 to 5.

This system have scored competitively in the task, with a $71.30\%$ Pearson correlation score. 
They attempted to run the same system without the machine translation step, which yielded a score that was significantly worse, $43.73\%$. 
This can be attributed to two factors. The first is the fact that the Arabic dataset is much smaller in comparison to the English dataset. 
This is corroborated by a similar drop in the score that is observed in the Spanish pairs’ track under the same conditions. 
The second is the fact that dense word representations, as obtained for a language such as English, can be inadequate for a morphologically rich language such as Arabic without incorporating these morphological features into the calculation of these word embeddings \cite{salama2018morphological}.
% Our methods do not use machine translation because it induces errors that propagate through the system and does not capture Arabic morphology.
Our methods do not use machine translation because it may produce incorrect translations especially when the system has to grasp the nuance of the text and take the context into account. 
This induces errors that propagate through the system and results in loosing information such as not capturing Arabic morphology.

% With the above forming the raison d'etre of our research, we next proceed into detailing the experimental setup of various deep learning models we have developed for tackling STS in Arabic.

%% file: sections/related/CL.tex
In the approach of computational linguistics, a sentence-based similarity measure could be defined by also counting for the syntactic structure of the sentence \cite{li2015calculation}. Authors of this paper divide the sentence into 3 components: subject, predicate, and object as the key components. However, they also include some modifier components which are attributive, adverbial, and complement. 
%Adverbs modifier was not included in their calculation as they showed that it could not be calculated.

Our approach combines the power of unsupervised learning through contextualized word embeddings and supervised deep pairwise fine-grained similarity network that outperforms state-of-the-art.

% There are text parsers that are utilized for tagging the sentence components. A recommended one is \textit{Stanford dependencies parser} \cite{de2006generating} which is an English language dependencies representation parser. Stanford parsers also have versions for other languages like Chinese, Arabic, French and German. There is a Chinese language special semantic and syntactic parser called Language Technology Platform (LTP) \cite{jun2006ltp}. It has been utilized in a related work by \cite{li2006sentence} among with HowNet \cite{dong2010hownet}. What is interesting in the syntactic-based algorithm proposed by \cite{li2006sentence} was counting for negativity and antonym by having the similarity value bounded by the range [-1.0, 1.0].

%% file: sections/data.tex
There are plenty of datasets available for STS tasks in English language, for example: SQuAD~\cite{DBLP:journals/corr/RajpurkarZLL16}, CoQA~\cite{DBLP:journals/corr/abs-1808-07042}, Quora Question Pairs (QQP) \footnote{https://data.quora.com/First-Quora-Dataset-Release-Question-Pairs} and Semantic Textual Similarity Benchmark \cite{cer2017semeval}. However, when it comes to the Arabic language there is a shortage and difficulty in finding such datasets. To overcome this issue we build our own Q2Q dataset for training and extract dialect test for evaluation. In addition, we collected large text corpus to train word and sentence embeddings. The following subsections provide more details on these datasets.

\subsection{Mawdoo3 Q2Q Dataset}
\input{sections/data/q2q.tex}

\subsection{Madar Dialect Q2Q Dataset}
\input{sections/data/dialectic.tex}

\subsection{Text Corpus}
\label{sec:textcorpus}
\input{sections/data/text_corpus.tex}

%% file: sections/data/q2q.tex
This dataset has been developed, annotated, and verified by the annotation team of \textit{Mawdoo3 Ltd}\footnote{https://ai.mawdoo3.com/nsurl-2019-task8} \cite{seelawi-EtAl:2019:NSURL} .
It consists of 3 fields, i.e. \textit{question1} containing the text for one of the question pairs, \textit{question2} containing the text of the second question, and \textit{label} which is either 1 if question1 and question2 have similar answer, or 0 if their answers are different. The collected questions are written (not spoken) factoid questions. For example:
\begin{center}
    \begin{RLtext}
        mn hw r'Is AlwlAyAt Almt.hdT Ala'mrykyT?
    \end{RLtext}
\end{center}
which translates to ``Who is the president of the United States of America?''.

% The total number of unique question pairs in the \textbf{training dataset} is $11,997$ with $6,600$ of them labeled as \textbf{0} and the remaining $5,397$ pairs are labeled as \textbf{1}. Before using the data to train the models, a pre-processing step consists of removing letter longations like \textarab{وو} and \textarab{يي}. 
The total number of unique question pairs in the \textit{training dataset} is 11,997. 6,600 of them are labeled as `0', and the remaining 5,397 pairs are `1'.

Before using the data to train the models, we run multiple pre-processing steps.
We remove all special characters and punctuation marks except Arabic punctuation marks such as commas and full stops.
We add space before and after the remaining punctuation marks so that they would be considered as tokens.
Our final pre-processing step is to remove letter elongations like \setcode{utf8} \<ww> (waw waw) and \setcode{utf8} \<yy> (ya ya).
% Please find arabic encoding table here https://en.wikipedia.org/wiki/ArabTeX
% \textarab{؟}
% Our final pre-processing step is to add space before and after the punctuation marks so that they would be considered as tokens.
The questions are relatively short, with most of the questions averaging between 4 to 10 words per question.

% \begin{figure}[tb]
% \centering
% \includegraphics[width=\columnwidth]{figures/hist_training.png}
% \caption{\label{fig:word-dist}
% Distribution of question lengths (word count) in Mawdoo3 Q2Q dataset (MSA).
% The figure on the left shows Question 1 histogram, and Question 2 on the right.
% % Number of words in question pairs.
% }
% \end{figure}

% \begin{figure}[htp]
%     \centering
%     \begin{subfigure}[t]{0.5\textwidth}
%         \centering
%         \includegraphics[height=1.2in]{figures/length_histogram.png}
%         \caption{\label{fig:word-dist} Number of words in question pairs.}
%     \end{subfigure}%
%     ~ 
%     \begin{subfigure}[t]{0.5\textwidth}
%         \centering
%         \includegraphics[height=1.2in]{figures/madar_data_t.png}
%     \caption{\label{fig:madar-dist} Number of words in Madar Q2Q dataset.}
%     \end{subfigure}
%     \caption{Caption place holder}
% \end{figure}

%% file: sections/data/dialectic.tex
% Previous work done by MADAR \cite{bouamor2018madar} to collect parallel corpora for Arabic dialects. Each data entry contains the MSAversion of a sentence, the dialect translation of that sentenceand the city's dialect. This dataset is based on tourism-related sentences, hence it is conversational in nature and comprises a lot of questions. This dataset contains parallel sentences for 24 cities.

MADAR dataset \cite{bouamor2018madar} is a parallel corpora of Arabic dialectic and MSA sentences.
Each entry in MADAR consists of a sentence in MSA, the dialectic translation of that sentence, and the city where the dialect is spoken.
This dataset contains dialects of 24 cities.
It is based on tourism-related sentences, hence it is conversational in nature and comprises a lot of questions. 

We use MADAR dataset to extract Q2Q pairs similar in format to the dataset described in the previous subsection (i.e., question1, question2, and label).
For each dialect, we first select the questions from MADAR dataset. 
Then, we label these question pairs with `1' to indicate that those questions are similar. 
Next, we generate negative samples by randomly matching a question from MSA with any other question within the same dialect. 
Finally, we balance the number of instances to make sure that all dialects have equal number of question pairs. 
Each dialect has 1,686 question pairs equally divided between similar pairs and the ones that are not. 
This will make a total of 40,464 pairs across all 24 dialects. 
We run the same pre-processing steps that we apply on Mawdoo3 Q2Q dataset.
% \begin{figure}[b]
% \centering
% \includegraphics[width=\columnwidth]{figures/hist_dialect.png}
% \caption{\label{fig:madar-dist} Distribution of question lengths  in Madar Q2Q dataset (MSA vs Dialectic).
% % The figure on the left shows Question 1 histogram, and Question 2 on the right.
% % Question1 and Question2 
% % Number of words in Madar Q2Q dataset.
% }
% \end{figure}

%% file: sections/data/text_corpus.tex
To train our embedding models (Sections \ref{sec:s2v} and \ref{sec:elmo}) we have collected a large dataset from three sources: (i) Tweets (ii) Arabic Wikipedia (iii) Mawdoo3 articles.

Twitter is a major data source for many NLP researchers.
This might be due to multiple reasons. Mainly the huge amount of data that can be collected and the diversity of the subjects that tweets can cover.

The Arabic content on the web may contain several local dialects (e.g., Egyptian, Moroccan, Gulf, Levantine dialects, etc..).
Arabic dialects can significantly vary from MSA.
Twitter contains a large amount of dialectic data. 
Naturally, tweets are not sequential corpus as each tweet is independent and not affiliated with previous nor preceding text.
Such data is important to build embedding models to represent dialects.
We have collected 68,400,000 unique tweets using the standard Twitter API. 

Wikipedia\footnote{www.wikipedia.com} is an encyclopedia of articles covering several topics.
\textit{Wikipedia.org} provides these articles for users to download.
We have downloaded the articles written in MSA, cleaned them, and segmented them into sentences based on three punctuation marks, namely: \{! ? .\}.
Segmenting the articles resulted in 4,600,000 sentences.

Mawdoo3\footnote{www.mawdoo3.com} is an encyclopedia providing articles in Arabic.
Similar to Wikipedia articles, we obtained Mawdoo3 articles and segmented them resulting in 2,800,000 sentences.

To clean the data that we have collected from the three mentioned sources we follow AraVec approach for text normalization~\cite{aravec}.
We combine all sentences in one dataset and use them to train our embedding models.

%% file: sections/models.tex
There are diverse approaches with multiple flavors that can be applied to tackle the problem of Q2Q similarity. In our research, we identify machine learning blocks that can be utilized to build an end-to-end prediction model.

\begin{figure}[tb]
 \centering
  \resizebox{0.9\linewidth}{!} {
%  \scalebox{.8}{%
% \includegraphics[width=\linewidth]{figures/building_blocks.png}
  \includegraphics[width=220pt]{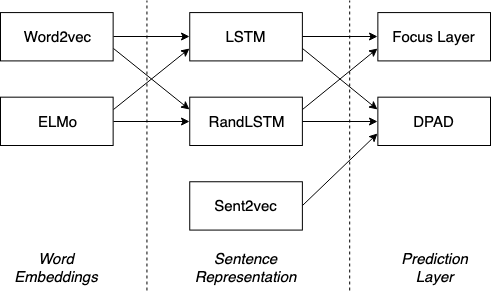}
}
 \caption{Building blocks of Q2Q network}
 \label{fig:network_illustration}
\end{figure}

Figure \ref{fig:network_illustration} illustrates the building blocks, which are grouped into 3 main categories:
\begin{enumerate}
  \item \textbf{Word Embeddings}: Converts a token into a vector representation that corresponds to its semantic meaning.
  \item \textbf{Sentence Representation}: Either built on top of word embeddings or generated dependently (i.e., Sent2vec).
  \item \textbf{Prediction Layer}: This layer consumes sentence embeddings from two different questions to output whether the questions are similar or not.
\end{enumerate}

In the subsections below, we describe each machine learning block and where it fits in the pipeline.

\subsection{Word2vec}
\label{sec:w2v}
\input{sections/models/word2vec.tex}

\subsection{ELMo}
\label{sec:elmo}
\input{sections/models/elmo.tex}

\subsection{Trainable LSTM}
\label{sec:trainablelstm}
\input{sections/models/trainable_lstm.tex}

\subsection{RandLSTM}
\label{sec:randlstm}
\input{sections/models/rand_lstm.tex}

\subsection{Sent2vec}
\label{sec:s2v}
\input{sections/models/sent2vec.tex}

\subsection{Focus Layer}
\label{sec:focaus}
\input{sections/models/focus.tex}

\subsection{Dot Product \& Absolute Distance (DPAD)}
\label{sec:dotdistance}
\input{sections/models/dotdistance.tex}

Figure \ref{fig:network_illustration} demonstrates how those building blocks are connected. It is worth noting that there is one invalid connection between Sent2vec and Focus layer. The reason for this is that Focus Layer compares array of vectors produced from each question. This array of vectors can be generated by LSTM cells, but Sent2vec can only compute one fixed-size thought vector per question.

There are nine different ways to connect those machine learning blocks. We build those models and train them with Mawdoo3 Q2Q dataset.

%% file: sections/models/word2vec.tex
Word2vec~\cite{cbow,word2vec} is a technique to learn continuous vector embeddings of words that represent the semantic meaning.
These embeddings are learned using one of two ways, either using a group of words as context to predict a target word, a method known as Continuous Bag of Words (CBOW), or using a word to predict context words, which is known as skip-gram.
By the end of the training phase, word vectors are projected in a high dimensional space such that similar words that share common contexts are located close to each other.
AraVec~\cite{aravec} is a widely-used pretrained Word2vec embedding model that is trained on Arabic tweets using CBOW method.
We use AraVec in our Word2vec experiments.

%% file: sections/models/elmo.tex
Despite the major advances in NLP brought about by dense vector representations of semantic units, it is not hard to see that it suffers from a few major drawbacks. One obvious shortcoming is that such representations fail to capture the different meanings that a given word might have (i.e. homonyms). For example, while the word ``can" has different meanings depending on the context, they all get the same semantic representation. It can be readily seen how this might affect the performance of the downstream applications which build on these word embeddings.

\begin{figure}[tb]
 \centering
  \resizebox{0.98\linewidth}{!} {
%  \scalebox{.8}{%
% \includegraphics[width=\linewidth]{figures/building_blocks.png}
  \includegraphics[width=220pt]{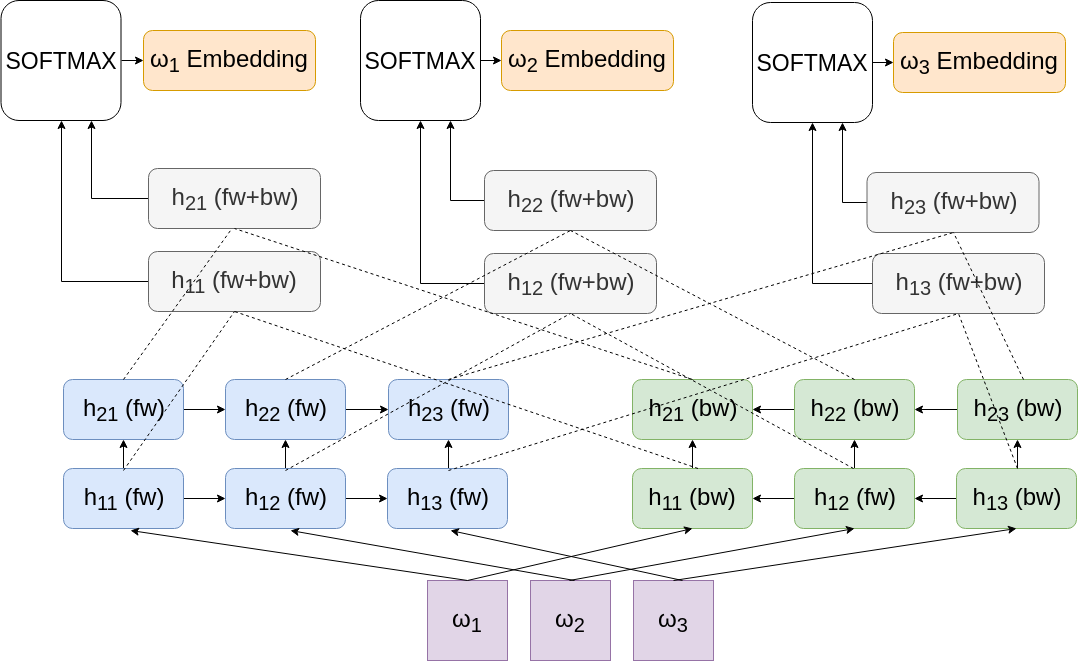}
}
 \caption{Illustration of the major components of an ELMO model}
 \label{fig:elmo_net}
\end{figure}

The past few years have witnessed an increased amount of research directed to addressing this shortcoming, such as in \cite{neelakantan2015efficient,choi2017context,devlin2018bert}. ELMo (Embeddings from Language Models) \cite{peters2018deep} representations are among the most successful of these. As shown in Figure \ref{fig:elmo_net}, these representations are obtained as a linear combination of the internal representations of the different layers of a bidirectional RNN language model. This can be thought of calculating the representations of any word in a given sentence, by conditioning on the whole sentence. The weighing of the contributions of each layer to the final representation, are fine-tuned per task. This allows the top layer machine learning algorithm to learn which of the signals from the various layers of the bidirectional language model are the most prevalent for its supervised learning task.

% \begin{figure}[tb]
%  \centering
%   \includegraphics[width=\linewidth]{figures/elmo2.png}
%   \caption{Illustration of the major components of an ELMO model}
% \end{figure}

%% file: sections/models/trainable_lstm.tex
This layer consumes a sequence of any word embedding type (i.e., Word2vec or ELMo) passing them into an LSTM \cite{hochreiter1997long} layer, one word embedding at each time step, and converts them into a thought vector. Thought vectors are a fixed sized array of numbers that represent the semantic meaning of the entire sentence, which is later consumed by a prediction layer. LSTM also computes an output vector for each time step which is compatible with another type of prediction layers.

%% file: sections/models/rand_lstm.tex
By using randomly initialized bidirectional LSTMs, to project pretrained word embeddings into a higher dimensional space, \cite{wieting2019no} were able to obtain very good results in sentence semantic similarity tasks. The LSTMs weights are sampled from the uniform distribution: 
\begin{equation}
\lbrack\cfrac{1}{\sqrt{-d}},\cfrac{1}{\sqrt{d}}\rbrack
\end{equation}

where $\boldsymbol{d}$ represents the latent size of the given LSTM. 
The size $\boldsymbol{d}$ should be substantially larger than the size of the embedding space of the word representations used as input. 
This is based on the empirical evidence presented by the authors in their benchmark tests. 
It also builds on the observation made by \cite{cover1965geometrical}, stating that for classification purposes, features that are projected into a high-dimensional space via some nonlinear method, has a better chance of being linearly separable than a corresponding lower dimensional projection, as long as the space is sparse in nature. 
The authors also note that the quality of the word representations used as the input, contributes the most to the performance obtained using such a configuration. 

%% file: sections/models/sent2vec.tex
Sent2vec~\cite{sent2vec} is an unsupervised model for sentence (not word) embeddings.
The model computes the average of vectors of all words and n-grams in a sentence.
% These word and n-gram vectors are trained to be averaged into sentence embedding.
% Although Sent2vec could be used to extract embeddings for words, it is trained to produce sentence embeddings.
Several studies have shown successful results produced by Sent2vec \cite{wieting-gimpel-2018-paranmt,AGARWAL2018922}.
% We trained our sent2vec model with the union of Twitter, Wikipedia, and Mawdoo3 data described in Section~\ref{sec:textcorpus}.
The model learns a context embedding $v_w$ and target embedding $u_w$ for each word $w$ in the vocabulary, with $h$ number of embedding dimensions.
The context words embeddings of a sentence, and its n-gram vectors, are averaged to get the sentence embeddings ($v_S$),
as illustrated in Equation \ref{eq:s2v}\cite{sent2vec}.

\begin{equation}
\label{eq:s2v}
v_S= \frac{1}{\left|R(S)\right|} \sum_{w\in R(S)} v_w~,
\end{equation}
where $R(S)$ is all the words and n-grams in sentence $S$.

%% file: sections/models/focus.tex
We adopt the work done by He \textit{et al.,}\cite{focus}. Focus layer receives pair of questions, tokenize each pair into words, then for each token (word), we look for its similar word in the other question of the pair. If a match is found, then that word is considered important by the algorithm. This algorithm will capture a word in a question or sentence pair with its context, compare that with each word and its context in the other question or sentence pair with its context.
% Intuitively, when a human is given a pair of questions to decide whether they are similar or not, the reader will look for words in both questions and decide on the pair similarity accordingly. This model has the same intuition, as the model is built to capture the word with its context, and compare it with other words and their contexts.
The model consists of four components as follows:
% \begin{enumerate}
%   \item Bidirectional LSTM for context modeling.
%   \item A pairwise word interaction modeling.
%   \item Similarity \textit{focus layer} for helping the model to identify word-level interactions in the sentence pair.
%   \item A deep CNN that detects similarity \textit{patterns} within the sentence pair as the final prediction step.
% \end{enumerate}
\subsubsection{Context modeling}
A bidirectional LSTM (Bi-LSTM) is applied to the input embedding of each question in the pair. This allows the context to be represented in word vectors.

\subsubsection{Pairwise word interaction modeling}
The model will compare each word in a question with all words in the other question. Then compute a similarity measure on the hidden states of the LSTM that represent each word.

\cite{focus} has defined a similarity metric that consists of \textit{cosine}, \textit{dot product}, and \textit{L$_2$ Euclidian distance} measures. 
All these measures are calculated between the hidden representations of each question of the pairs. 
The following Equation (Eq.~\ref{eq:eq1}) represents the similarity metric mathematically.
% We can represent this mathematically as follows :
\begin{equation}
\label{eq:eq1}
\text{sim}(\mathbf{h_1} , \mathbf{h_2}) = \left \{\cos(\mathbf{h_1} , \mathbf{h_2}), \left\|\mathbf{h_1}, \mathbf{h_2}\right\|, \inp{\mathbf{h_1}}{\mathbf{h_2}} \right \},
\end{equation}
%  $$\text{sim}(\mathbf h_1 , \mathbf h_2) = \{\cos(\mathbf h_1 , \mathbf h_2), \left\|\mathbf{h_1}, \mathbf{h_2}\right\|, \( \inp{\mathbf h_1}{\mathbf h_2} \) \}$$ 
where:
\begin{itemize}
 \item $\mathbf{h_1}$ and $\mathbf{h_2}$ are the hidden states of the Bi-LSTM for question 1 and question 2, respectively .
 \item $\left\|\mathbf{h_1}, \mathbf{h_2}\right\|$ 
 is L$_2$ Euclidian distance between $\mathbf h_1$ and $\mathbf h_2$
 \item $\inp{\mathbf{h_1}}{\mathbf{
 h_2}}$ is the inner (dot) product of $\mathbf h_1$ and $\mathbf h_2$
\end{itemize}
The output of this step is called \textit{simCube} (or Similarity cube). A more detailed discussion on how the \textit{simCube} is built can be found in \cite{focus}. Note that the \textit{simCube} is built around the word contexts not the words themselves.
\subsubsection{Focus layer}
The core of this model is the focus layer that builds on top of the pairwise word interaction model and introduces re-weighting mechanism for the \textit{simCube} model.

Since two types of similarity measures are used in Equation \ref{eq:eq1}, namely, cosine and L$_2$ distance measures, the focus layer output (called \textit{focusCube}) is computed over the two types of similarities. 
The computation aims to maximize the similarity for all word interactions computed in the \textit{simCube}. 
The focusCube represents the final computation of the similarity score between words in the question pair.
\subsubsection{ResNet}
FocusCube converts Q2Q problem into pattern recognition problem, as such; a convNet of 19 layers has been used, followed by 2 fully connected layers, and a log-softmax layer as the final computation of the similarity score.

%% file: sections/models/dotdistance.tex
Similar to Focus layer, DPAD is one of the options for prediction layer at which we decide the similarity score and has been widely used in the literature~\cite{sent2vec,shao2017hcti}. 
To predict the similarity between a pair of questions, first, we extract each question's thought vector using one of the sentence embedding methods.
Then we apply two operations between the two vectors: (i) The absolute element-wise difference  (ii) The element-wise multiplication.
These operations result in two vectors. We concatenate them into one vector and feed it into logistic regression model, which predicts whether the two sentences are similar.

%% file: sections/experimental.tex
In this section, we demonstrate the effectiveness of our proposed approach.
To conduct our experiments we have assembled models from the machine learning blocks discussed in Section~\ref{sec:models}.
We compare these models to each other and to state-of-the-art baseline (described below).
We claim that using Focus layer with ELMo embedding to predict Q2Q similarity boosts up the prediction accuracy and outperforms state-of-the-art approaches.

All experiments have been conducted on an Nvidia Tesla V100 GPU machine, with Ubuntu 16.04 as the operating system. All models, except for the Focus layer, are implemented with TensorFlow 1.12 \cite{abadi2016tensorflow} which is implemented using Pytorch 1.1 \cite{paszke2017pytorch}.

% we have obtained AraVec (a word2vec model trained with Arabic Tweets) and trained Sent2vec and ELMo models using Twitter, Wikipidea and Mawdoo3 data.

\subsection{Baseline}
\input{sections/models/weighted_embd.tex}

\subsection{Trained Models}
\input{sections/experimental/training.tex}

\subsection{Results and Discussion}
\input{sections/experimental/results.tex}

%% file: sections/models/weighted_embd.tex
Several studies have shown that using averaged word embedding vectors outperforms sophisticated LSTM-based sentence embeddings
~\cite{arora2017,sent2vec}.
Arora et al.~\cite{arora2017} have proposed an approach called smooth inverse frequency (SIF) for generating sentence embeddings.
We briefly describe it here.
First, the weighted average of pre-trained word embedding vectors of the input sentence is calculated.
The weights are computed using the frequency of words in training corpus.
Frequent words get smaller weights than less frequent ones.
% Finally, the first component of the average vectors is removed using SVD.
Then the projection of the average vectors on their first singular vector is removed using SVD.

% Here we briefly describe the approach. 
This approach has been widely used and has proven to be a strong baseline.
% So, we use it as a baseline.
% We generate sentence embeddings.
We compute sentence embeddings for questions using this approach.
We use AraVec as the pre-trained word vectors.
Then we compute \textit{DPAD} (as described in Section~\ref{sec:dotdistance}) to predict whether two questions are similar.

% Weighted word emb. sum with principle component removal

% word vectors trained using the GenSim word2vec implementation

% Surprisingly, for constructing sentence embeddings, naively using averaged word vectors outperforms LSTMs for weighted averaging

% It is a weighted average of word embeddings in the window, with smaller weights for more frequent words (reminiscent of tf-idf).

% proposed methods in which sentences are represented as a weighted
% average of fixed (pre-trained) word vectors

%% file: sections/experimental/training.tex
% Discuss the following points:
% \begin{enumerate}
%     \item Training Sent2vec (dataset used, hyperparameters, Hardware and time needed)
%     \item Training ELMo (dataset used, hyperparameters, Hardware and time needed)
%     \item Training our models  (dataset used, hyperparameters (RandLSTM, focus, dot trainable LSTM ... etc.), Hardware and time needed) 
%     \item Baseline (copy section in models here)
% \end{enumerate}

\subsubsection{Baseline}
To produce vectors using the baseline method, we use SIF implementation\footnote{https://github.com/PrincetonML/SIF}.
The weights of words are computed using the number of word occurancies in the training dataset.
Following the authors approach \cite{arora2017}, we set the weighting parameter to 0.001 and remove only one principle component.
As mentioned above, we use AraVec pre-trained word embeddings.

\subsubsection{ELMo}
ELMo weights are trained on the text corpus discussed in Section \ref{sec:textcorpus}. Then, for fine-tuning, we train ELMo connected to DPAD with Mawdoo3 Q2Q dataset. 
We set hyper-parameters configuration as shown in table \ref{tab:hyperparams}. 
We have trained 64 sentences per batch, and each training epoch took around three minutes.

It is worth noting that BERT \cite{devlin2018bert} is not part of our experiments. We can not train BERT with ``Next Sentence Prediction'' task because we do not have a continuous dialectic corpus. Tweets on the other hand is sufficient to train ELMo.

\subsubsection{Sent2vec}
To train our Sent2vec embedding model we use the package\footnote{https://github.com/epfml/sent2vec} that is built on Facebook's FastText library.
We train the Sent2vec model of 300 dimensions using Negative Sampling loss function with negative samples set to 25.
We did not specify a limit for the vocabulary size (i.e., maxVocabSize=None), however, we set the minimum number of a word occurrences (minCount) to 100 to eliminate uncommon typos in the dataset.
The maximum length of word ngram is 2 and the number of epochs is 5. Similar to ELMo, Sent2vec is also trained on the text corpus collected from Mawdoo3, Wikipedia, and Twitter.
\subsubsection{Focus layer}
It is used in two settings: (i) Word2vec (ii) ELMo
% for word embeddings.
We set the hyper-parameter \textit{hidden-units} to 256 for Word2vec, and 512 for ELMo.
% For the first setting, the hyper-parameter \textit{hidden-units} is 256 while the , with 
The other hyper-parameters are the same for both as shown in Table \ref{tab:hyperparams}. We have trained the Focus layer for 20 epochs. On average, each epoch took around 90 seconds.

%% file: sections/experimental/results.tex
We compare the baseline with the nine different combinations composed from building blocks discussed in section \ref{sec:models}. All models are trained with Mawdoo3 Q2Q training set and then evaluated against two different testing sets: Mawdoo3 Q2Q test set, and Madar Dialect Q2Q dataset. We report our F1 score results in table \ref{tab:results}. 

% Baseline
We first report the baseline, which achieves relatively high score on MSA test set with 0.72 F1 score but fails to generalize on dialectic test set scoring 0.42 F1 score, due to the fact that Word2vec works on word level and can not cope with Arabic language morphology. It can also be attributed to the small number of weights used in this approach.

% sent2vec + DPAD
The second trial is the Sent2vec combined with DPAD which is able to break the baseline on MSA (0.76 F1 score) but also fails to generalize on dialect dataset (0.36 F1 score) because, as in the baseline, Sent2vec neither emphasis on the morphology of the word as in ELMo nor has the number of learnable parameters that ELMo has.
% of the same factors mentioned for the baseline. 

% Word2vec + TrainableLSTM + DPAD
Next we try Word2vec topped with LSTM layer, drastically increase the number of learnable weights, and DPAD prediction layer. 
The trainable LSTM layer boosts prediction for both MSA achieving 0.81 F1 score and 0.66 F1 score on dialectic test set. Formulating a thought vector from word embeddings is able to capture sentence semantics better than previous approaches. But this approach still suffers to generalize on dialects as the gap between both test sets is large, around 0.15 F1 score difference. 

% Word2vec + TrainableLSTM + Focus
We test the impact of the focus layer by using it as a prediction layer instead of DPAD. As expected, adding a pairwise word similarity sub-network improves the performance of our model on MSA (0.84 F1 score). It is able to overcome word order transformations such as topicalization by finding patterns on word by word similarity scores to compute the final prediction. It is able to generalize better on dialectic test achieving 0.70 F1 score.

% ELMo
Next, we examine the impact of ELMo contextualized word embeddings by using it instead of Word2vec in all previous experiments. ELMo consistently improves accuracy in all combinations, which proves that our ELMo model, trained on Tweets, Arabic Wikipedia and Mawdoo3 articles, is better than AraVec and is able to capture semantic meaning more accurately. Arabic language is highly derivational language mutating many morphological variations \cite{habash2018unified} for each word, ELMo is able to capture sub-word embeddings such as root words, suffixes and prefixes generalized enough to project words accurately in all of its variations. In addition, ELMo can differentiate homonyms by projecting their embeddings based on their context.

% RAND LSTM Discussion
Finally, we inspect the effect of replacing a trainable LSTM with random fixed LSTM. Trainable LSTM models have higher capacity to learn from the training dataset, hence it is expected to get higher scores than RandLSTM. But, it is interesting to note that when it is combined with focus layer, the gap between MSA and Dialect F1 scores is almost eliminated. We think that having fixed weighted LSTM acts as regularization mechanism for the trainable weights of the focus layer.

% Best Models
Experiments show that the best model for MSA Q2Q dataset is ELMo with trainable LSTM and DPAD, achieving an F1 score of 0.93. Unfortunately, this approach is not able to perform well on Madar Dialectic Q2Q as it is limited with thought vector. On ther other hand, replacing DPAD with focus layer makes the solution robust across all test datasets, by getting 0.90 F1 on MSA and the highest score of 0.82 F1 score on Madar Dialectic Q2Q.

% Dialect discussion
\textcolor{black}{Throughout} our experiments, it is clear to notice that focus layer is consistently making better predictions on Madar Dialectic Q2Q than DPAD. We calculate the average F1 difference between MSA and Dialectic for models using focus layer (0.073), then for models using DPAD (0.233). Having a deeper look into both test datasets, we notice that there are three main differences:

\textbf{Sentence Lengths:} Dialectic sentences are shorter in length. For example, MSA question:

\setarab % choose the language specific conventions
% \vocalize % switch diacritics for short vowels on
% \transtrue % additionally switch on the transliteration
\arabtrue % print arabic text ... is on by default anyway
\vspace{-1mm}
\begin{center}
    \begin{RLtext}
        'anA b.hAjT 'il_A t_dkrT?
    \end{RLtext}
\end{center}
\vspace{-1mm}

% \begin{otherlanguage}{arabic}
% أنا بحاجة إلى تذكرة ؟
% \end{otherlanguage}

% \setarab
% \novocalize
%  \arabtrue
% \RL{anA b.hAj/^gT Aly/I* t_dkrT?}

\begin{figure}[!htb]
\centering
\includegraphics[width=\columnwidth]{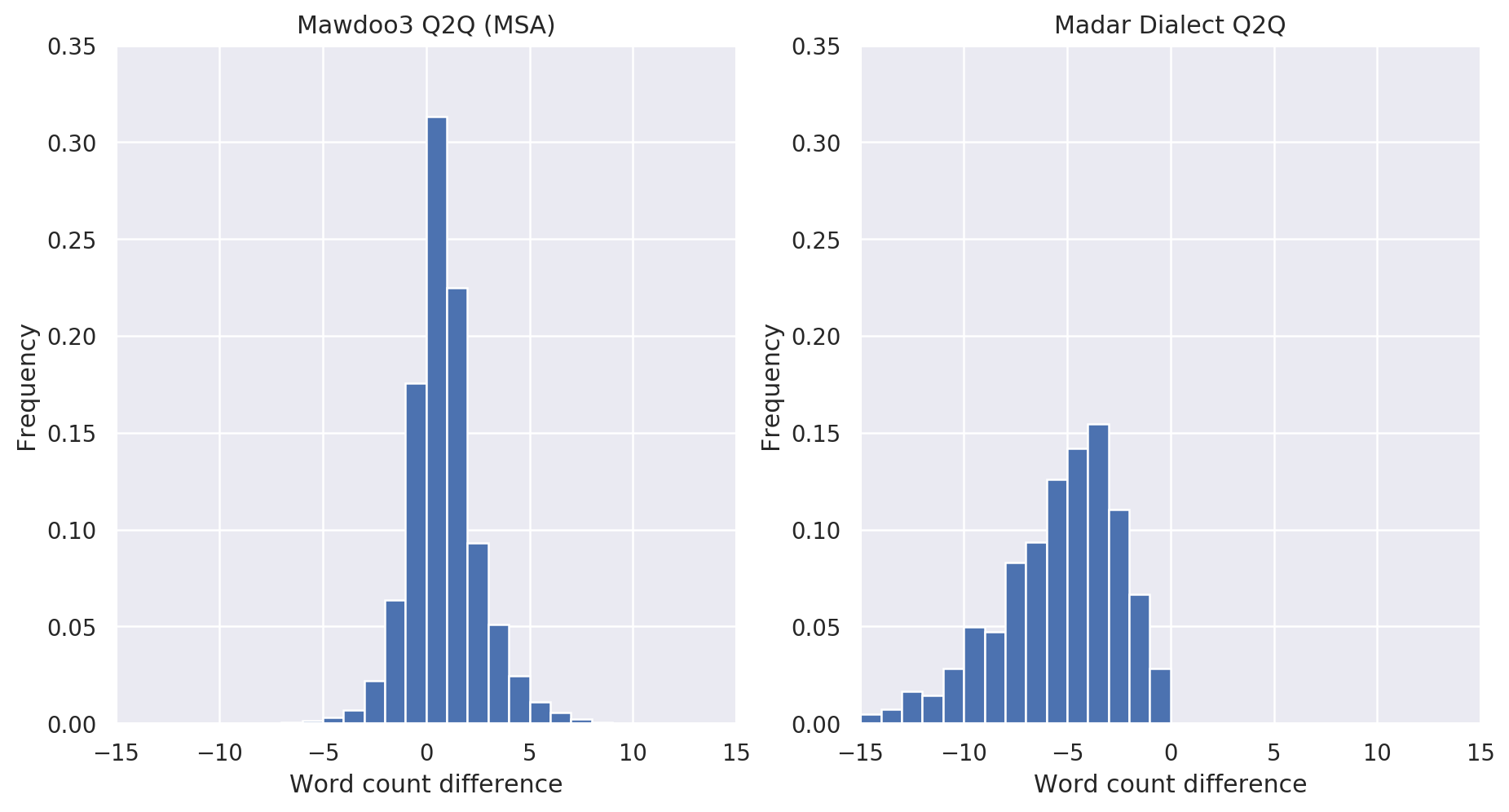}
\caption{\label{fig:hist_pair_diff}
Distribution of question lengths (word count) difference between a pair in Mawdoo3 Q2Q dataset (MSA) and Madar Q2Q dataset (dialect).
The figure on the left shows that the difference between question lengths of a pair in Mawdoo3 Q2Q dataset is small, varying 5 words at most, while the figure on the right, shows the difference between question lengths of Madar Q2Q dataset is significant, with the first question (MSA) has higher number of words than the second question (dialectic).
% Number of words in question pairs.
}
\end{figure}

% \setcode{utf8}
% \<a'nA b.hAj/^gT Aly/I* t_dkrT?>
% \< بحاجة إلى تذكرة>
% \begin{arab}[voc]
% أنا بحاجة إلى تذكرة ؟
% \end{arab}
% \setcode{utf8} (\<ش>)
% "\textarabic{أنا بحاجة إلى تذكرة؟} "
% \begin{center}
%     \begin{RLtext}
%         a'nA b.hAj/^gT Aly/I* t_dkrT?
%     \end{RLtext}
% \end{center}

which means ``Do I need a ticket?'' contains four tokens, while the dialect version:
\vspace{-1mm}
\begin{center}
    \begin{RLtext}
        b.htAj t_dkrT?
    \end{RLtext}
\end{center}
\vspace{-1mm}
% "\textarabic{بحتاج تذكرة؟}"
% \begin{center}
%     \begin{RLtext}
%         b.htAj/^g t_dkrT?
%     \end{RLtext}
% \end{center}
has only two tokens. Hence, our model has to be capable of comparing long sentence (MSA) with shorter one (dialectic). Figure \ref{fig:hist_pair_diff} shows the difference of question pair lengths in each dataset. The figure on the left is concerned with the difference of lengths between question 1 and question 2 in Mawdoo3 Q2Q dataset, as depicted, the question pairs are close in word count (length). On the other hand, it is clearly noticeable from the right-hand figure, the lengths of Madar Q2Q dataset question pairs have higher variance, yet the Focus Layer is still able to perform very well on this case.

\textbf{Word Order:} Written Questions are deep structured in which ambiguities do not exist and the semantic interpretation of a question is clear \cite{chomsky1969deep}. While dialectic questions are surface structured in which the sentence appears in speech. This means that our model has to overcome word order transformation to make a better predictions on dialect dataset. 

\textbf{Domain:} MSA dataset is a collection of factoid questions \textcolor{black}{whereas} dialect dataset is conversational.

We attribute the success of focus layer by its ability to overcome sentence lengths and word order phenomena. Focus layer compares all combination of word representation instead of thought vectors, which enables it to search for similarities between words despite length and word order differences.

%%%%%%%%%%%%%%%%%%%%%%%%%%% TABLE %%%%%%%%%%%%%%%%%%%%%%%%%%%%%%%

\begin{table}[!htb]
\centering
\caption{F1 scores on the test sets (MSA and Dialectic Arabic). The training set is MSA. Best results are in bold. The combination of ELMo and FocusLayer boosts up the performance on Dialect, yielding better generalization}
\resizebox{0.99\linewidth}{!} {
%  \scalebox{.99}{%
\begin{tabular}{|l|l|l|}
\hline
\centering \textbf{Model} & \textbf{MSA F1} & \textbf{Dialect F1} \\
\hline
Baseline (Weighted Word2vec + DPAD) & 0.72 & 0.42 \\
\hline
Sent2vec + DPAD & 0.76 & 0.36 \\
\hline
Word2vec + TrainableLSTM + DPAD & 0.81 & 0.66 \\
Word2vec + RandLSTM + DPAD & 0.62 & 0.44 \\
Word2vec + TrainableLSTM + FocusLayer & 0.84 & 0.70 \\
Word2vec + RandLSTM + FocusLayer & 0.72 & 0.69 \\
\hline
ELMo + TrainableLSTM + DPAD & \textbf{0.93} & 0.69 \\
ELMo + RandLSTM + DPAD & 0.83 & 0.47 \\
ELMo + TrainableLSTM + FocusLayer & 0.90 & \textbf{0.82} \\
% ELMo + NoLSTM + FocusLayer & 0.79 & 0.76 \\
ELMo + RandLSTM + FocusLayer & 0.75 & 0.71 \\
\hline
\end{tabular}
}
\label{tab:results}
\end{table}
%%%%%%%%%%%%%%%%%% hyperparams%%%%%%%%%%%%%%%%%%%%%
\begin{table}
\centering
\caption{Summary of Training Hyper-parameters for each model}
\resizebox{0.99\linewidth}{!} {
%  \scalebox{.99}{%
\begin{tabular}{|l|l|}
\hline
\centering \textbf{Model} & \textbf{hyper-params} \\
\hline
ELMo & \{bidirectional: true, batch-size: 128, \\
& lstm: \{cell-clip: 3, proj-clip: 3, dim: 4096,projection-dim: 512, \\
& use-skip-connections: true, n-layers: 2\}, \\
& n-negative-samples-batch: 8192, n-tokens-vocab: 6692, \\
& dropout: 0.1, n-epochs: 10, all-clip-norm\_val: 10.0, \\
& char-cnn: \{embedding: \{dim: 16\}\},max-characters-per-token: 50, \\
& activation: relu, n-highway: 2, n-characters: 262 \}\} \\
\hline
FocusLayer & \{regularization: 0.00005, \\
& optimizer: adam, batch-size: 128, learn-rate: 0.00005, \\
& hidden-units: 512, res-layers: 19, res-fmaps: 32, epsilon: 1e-8\} \\
\hline
Sent2vec & \{mincount = 100, dim = 300, loss = neg(25) \}\\
\hline
RandSent & \{out-dim:4096,max-seq-len:96,pooling:max,n-layers:1,projection:same \} \\
\hline
\end{tabular}
}
\label{tab:hyperparams}
\end{table}

%% file: sections/conclusion.tex
Our major contribution is further generalizing Q2Q models trained on standard language into supporting different variations and domains of the same language. The approach combines ELMo, a character level contextualized word representation to overcome the high morphological nature of the Arabic language, and a focus layer, to handle the word order transformations that is induced by dialectic formatting or by the nature of the language itself.

We also provide the largest MSA Q2Q dataset and a procedure to generate Dialectic Q2Q dataset utilizing the MADAR parallel corpora for Arabic dialects.

%Hussein
As a future work, we would like to use computational linguistics approach as a text pre-processing layer to the question pair. For example, using lemmatization, morphological segmentation, syntactical parser as well as synonym expansion to the question pair.